\documentclass[10pt,twocolumn]{article}

\usepackage[utf8]{inputenc}
\usepackage[T1]{fontenc}
\usepackage[margin=2cm]{geometry}
\usepackage{amsmath,amssymb,amsfonts,amsthm}
\usepackage{graphicx}
\usepackage{booktabs}
\usepackage{natbib}
\usepackage{hyperref}
\usepackage{xcolor}
\usepackage{enumitem}
\usepackage{subcaption}
\usepackage{multirow}
\usepackage{tabularx}
\usepackage{algorithm}
\usepackage{algorithmic}
\usepackage{microtype}

\newtheorem{theorem}{Theorem}

\newtheorem{definition}{Definition}

\definecolor{mosaicblue}{RGB}{31,119,180}
\definecolor{mosaicred}{RGB}{214,39,40}

\hypersetup{
  colorlinks=true,
  linkcolor=mosaicblue,
  citecolor=mosaicblue,
  urlcolor=mosaicblue,
}

\title{%
  Accurate and Efficient Long-Term Memory for LLM Agents%
}

\author{
  Zicheng Zhao$^{1,\dagger}$,\quad
  Xinyang Guo$^{1,\dagger}$,\quad
  Luyao Lv$^{2}$,\quad
  Menghan Wang$^{2}$,\\[4pt]
  Ming Li$^{1,*}$,\quad
  Shuaicheng Li$^{3,*}$\\[8pt]
  {\small $^{1}$Central China Research Institute for AI Technology}\\
  {\small $^{2}$Guangdong Provincial Key Laboratory of Interdisciplinary Research}\\
  {\small and Application for Data Science, BNU-HKBU United International College, Zhuhai}\\
  {\small $^{3}$Department of Computer Science, City University of Hong Kong}\\[4pt]
  {\small $^{\dagger}$Equal contribution \quad $^{*}$Corresponding authors}
}

\date{}

\begin{document}
\maketitle

\begin{abstract}
LLM agents augmented with persistent memory can recall past interactions, but existing systems suffer from two limitations: flat, unstructured storage loses relational context needed for multi-hop and temporal reasoning, and reliance on expensive LLM-based classification makes them impractical for latency-sensitive deployment. Without mechanisms to validate new information against stored knowledge, these systems silently accumulate contradictions. We present MOSAIC (Memory-Organized Structured Agent for Information Collection), a structured, conflict-aware long-term memory framework for LLM agents that is substantially more accurate and efficient. MOSAIC introduces three key capabilities: (1) entity-typed graph storage with semantic classification preserving relational structure across events, personas, and relationships, enabling multi-hop and temporal reasoning over conversation history; (2) hash-accelerated dual-path retrieval replacing LLM-based classification with locality-sensitive hashing, achieving near-instantaneous lookup with negligible accuracy loss; and (3) active conflict detection at save time that cross-references new information against existing graph neighbors, triggering updates or deletions for contradictory entries. Evaluated on LoCoMo (long-conversation QA), HaluMem, and a novel clinical-guideline error compounding test, MOSAIC achieves 89.35\% accuracy on LoCoMo (+27.21 pp over the best baseline), best HaluMem-Medium extraction F1 (86.77\%) and HaluMem-Long extraction F1 (85.84\%), best QA correctness on both Medium and Long (73.10\%, 70.75\%), and detects 66\% of injected factual conflicts-4.7 times higher than the best baseline (14\%)-while hash-accelerated retrieval keeps average search latency at 0.58\,s per question.
\end{abstract}


\section{Introduction}
\label{sec:intro}

The proliferation of memory-augmented LLM agents has created an implicit
assumption: that an agent which remembers is an agent that reasons correctly
over its memories. Systems such as MemGPT~\citep{packer2023memgpt},
Mem0~\citep{chhikara2025mem0}, A-Mem~\citep{xu2025mem}, and
MemOS~\citep{li2025memos} have demonstrated impressive retention of user
preferences, factual history, and contextual nuance across sessions. Yet a
fundamental limitation persists: these systems treat memory as
\emph{append-only storage}, ingesting new information without validating it
against existing knowledge, and thus silently accumulating contradictions
that compound across retrieval and reasoning stages.

This \emph{error compounding} problem is particularly acute in
safety-critical domains. In clinical settings, silently storing that a
patient's target blood pressure is $<$160/100~mmHg when guidelines specify
$<$130/80~mmHg propagates into incorrect risk stratification and potentially
harmful prescribing~\citep{yaraghi2024generative}. In legal and financial
contexts, contradictory facts feed contradictory retrievals, which feed
contradictory downstream decisions. The pipeline structure of modern memory
systems---from extraction to storage to retrieval to
reasoning---means that a single undetected error at ingestion can
exponentially amplify through subsequent stages.

A growing body of evidence reveals a deeper disconnect between \emph{recall}
and \emph{task completion}: memory-augmented agents routinely terminate
conversations prematurely, miss required information, or drift into tangential
topics when no explicit mechanism enforces systematic
coverage~\citep{xi2025rise, wang2024survey}. Current agent
frameworks provide two capabilities: \emph{action selection} (choosing what
to do next) and \emph{memory} (storing what was learned). But structured
information-gathering tasks require a third: \emph{task-state tracking}---a
persistent, queryable representation of which required information units have
been obtained, which remain outstanding, and what dependencies constrain the
order in which they should be sought. Without this, an agent cannot
distinguish between ``I have enough information to stop'' and ``I have run
out of things that are easy to ask.''

This distinction is not merely an engineering detail. In clinical intake, a
missed medication history propagates into incorrect risk stratification and
potentially harmful prescribing~\citep{singhal2023large}. In insurance
claims processing, a missing liability determination invalidates downstream
adjudication. In technical support, an undiagnosed hardware failure leads to
repeated software-only fixes. The common structure across these domains is a
\emph{partially observable information space} with \emph{prerequisite
constraints} and \emph{coverage requirements} that must be systematically
satisfied through multi-turn interaction.

We trace these failures to three structural deficiencies in existing memory
architectures: \emph{flat storage}, where memories are kept as unstructured
text chunks or key-value pairs without typed relationships, making
cross-referencing impossible; \emph{passive ingestion}, where new information
is stored unconditionally without conflict checking against existing
memories; and \emph{monolithic retrieval}, where semantic search over flat
stores cannot exploit relational structure for multi-hop or temporal
reasoning.

A deeper question is \emph{why graphs} rather than simpler task
representations such as flat checklists, priority queues, or plan templates.
The answer lies in a structural property of real information-gathering tasks:
\emph{local dependency}. Whether a particular information entity needs to be
queried, re-queried, or deprioritized depends not on the global state of
the entire conversation but on the states of a small set of semantically
or logically related entities---its \emph{graph neighbors}. Knowing a
patient's creatinine level affects the importance of asking about nephropathy
but not about smoking history. Confirming the date of an accident changes the
relevance of weather-condition queries but not vehicle-make queries. This
locality is precisely what graph structure encodes and what flat
representations discard.

We introduce \textbf{MOSAIC}, a framework that addresses these deficiencies
through three mechanisms. First, MOSAIC organizes memories as a \emph{typed
entity graph} with semantic classification into events, personas, and
relationships, preserving relational structure that enables graph-based
reasoning. Second, MOSAIC performs \emph{active conflict detection at save
time}: when new information is ingested, it is cross-referenced against the
existing memory graph; conflicts trigger warnings and may be resolved through
updates or deletions. Third, MOSAIC employs \emph{hash-accelerated retrieval}
that achieves near-instantaneous memory lookup while maintaining accuracy
comparable to LLM-based classification.

We formalize the key insight underlying MOSAIC's graph structure as
\emph{neighbor-conditioned stability} (NCS): a node's importance score,
belief state, and priority require re-evaluation only when at least one of
its graph neighbors has changed state. When the local neighborhood is
unchanged, the node's state is guaranteed stable. This principle is closely
analogous to Bellman's principle of optimality in dynamic
programming~\citep{bellman1966dynamic}, which decomposes a global
optimization into local value updates that propagate through a
state-transition graph. Just as dynamic programming avoids redundant
global recomputation by exploiting local structure, MOSAIC avoids redundant
re-evaluation of entity priorities by exploiting the dependency and
association structure encoded in the memory graph.

We evaluate MOSAIC on three complementary benchmarks: LoCoMo~\citep{maharana2024evaluating}, which tests long-conversation memory QA across 1,540 question--answer pairs; HaluMem~\citep{chen2025halumem}, which separately evaluates memory extraction, updating, and question answering with hallucination detection; and a novel error compounding test in which 50 factual conflicts are manually injected into hypertension clinical guidelines. Across these benchmarks, MOSAIC achieves 89.35\% overall accuracy on LoCoMo (+27.21~pp over the best baseline Mem0), best HaluMem-Medium extraction F1 (86.77\%), best HaluMem-Long extraction F1 (85.84\%), best HaluMem QA correctness on both Medium and Long (73.10\%, 70.75\%), detects 66\% of injected factual conflicts---4.7$\times$ higher than the best baseline (14\%)---and keeps average HaluMem-Medium search latency at 0.58~s per question.

\section{Related Work}
\label{sec:related}

\subsection{Memory-Augmented LLM Agents}

The development of persistent memory for LLM agents has progressed rapidly.
MemGPT~\citep{packer2023memgpt} introduces OS-inspired virtual memory
management, maintaining a tiered memory hierarchy (main context, archival
storage, recall storage) with explicit memory management functions.
Mem0~\citep{chhikara2025mem0} provides multi-level persistent memory with adaptive
personalization across user, session, and agent levels.
A-Mem~\citep{xu2025mem} implements adaptive memory with automatic storage,
retrieval, and summarization, dynamically adjusting memory granularity.
MemOS~\citep{li2025memos} unifies memory operations through an
operating-system abstraction layer.
Zep~\citep{rasmussen2025zep} provides production-grade memory with temporal awareness
for AI assistants.

A comprehensive survey by \citet{zhang2025survey} categorizes memory
mechanisms along dimensions of storage format, retrieval strategy, and
management policy. Despite this diversity, all existing systems share a common
limitation: they focus on \emph{retention}---ensuring that information, once
stored, can be retrieved accurately---but lack mechanisms for \emph{conflict
detection} or \emph{structured organization} that would prevent contradictory
information from being stored in the first place.

\subsection{Agent Frameworks and Planning}

The broader landscape of LLM agent architectures provides context for
MOSAIC's design choices. ReAct~\citep{yao2022react} interleaves reasoning
and acting, using chain-of-thought prompting to decide actions step by step.
Reflexion~\citep{shinn2023reflexion} adds verbal reinforcement learning
through self-reflective feedback loops. Plan-and-Solve~\citep{wang2023plan}
improves zero-shot reasoning through explicit plan generation before
execution. These frameworks demonstrate sophisticated planning and reasoning
capabilities but lack persistent, structured memory for cross-session
information management~\citep{xi2025rise, wang2024survey}.

Tool-augmented approaches~\citep{schick2023toolformer} extend agents with
external capabilities but do not address the fundamental memory organization
problem. Chain-of-thought prompting~\citep{wei2022chain} improves reasoning
within a single context window but does not support long-term memory across
sessions.

\subsection{Knowledge Graphs and Structured Memory}

Several systems organize agent memory as knowledge
graphs~\citep{pan2024unifying}. Mem0-Graph extends Mem0 with graph
structure but does not perform conflict checking. GraphRAG~\citep{edge2024local}
uses graph-based retrieval for query-focused summarization, demonstrating the
benefit of graph structure for information retrieval. However, these systems
focus on retrieval quality rather than ingestion-time validation.

MOSAIC's entity graph differs from general knowledge graphs in its focus on
typed entities (event, persona, relationship) with explicit conflict detection
at ingestion time. The dual-graph formulation---separating prerequisite
dependencies from semantic associations---provides both logical validity and
conversational coherence, properties that single-graph approaches cannot
simultaneously optimize.

\subsection{Memory Evaluation Benchmarks}

LoCoMo~\citep{maharana2024evaluating} evaluates long-conversation memory through
multi-type QA spanning single-hop, multi-hop, temporal, and open-domain
questions. HaluMem~\citep{chen2025halumem} introduces hallucination-aware
evaluation of the memory pipeline, separately assessing extraction, updating,
and question answering stages. LLM-as-Judge
approaches~\citep{zheng2023judging} supplement reference-based evaluation. Our
error compounding test complements these by specifically targeting conflict
detection, a capability not evaluated by existing benchmarks.

\subsection{Hallucination and Factual Consistency}

The hallucination problem in LLMs has been extensively
studied~\citep{huang2025survey, ji2023survey}. Fact
verification systems~\citep{thorne2018fever} and fine-grained factual
evaluation~\citep{min2023factscore} address hallucination at generation time.
MOSAIC addresses a complementary problem: preventing hallucinated or
contradictory information from being stored in persistent memory, where it
can compound across future retrieval and reasoning steps.

\section{Method}
\label{sec:method}

\begin{figure*}[!t]
  \centering
  \includegraphics[width=0.95\textwidth]{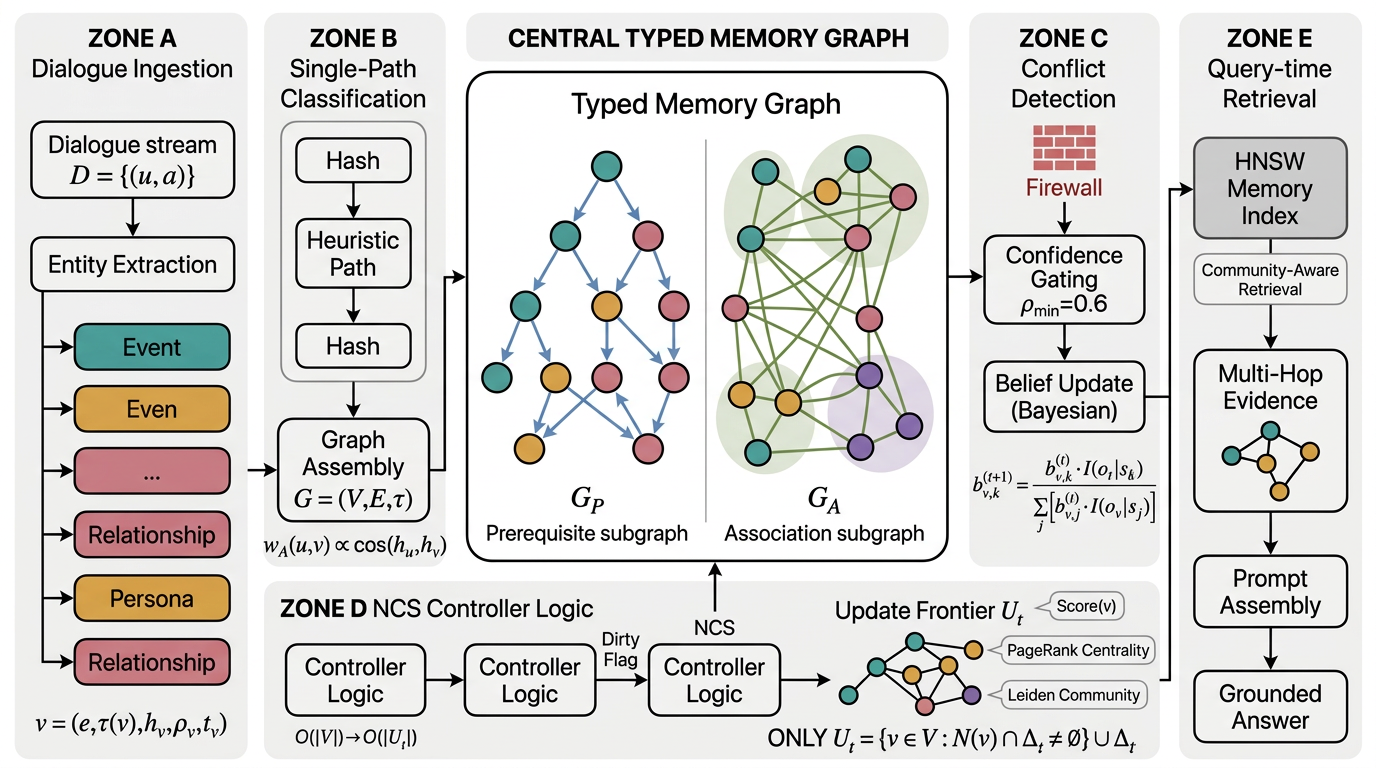}
  \caption{\textbf{MOSAIC system architecture.} Dialogue is processed
  through entity extraction and dual-path classification via LLM or
  locality-sensitive hashing. Entities populate a typed memory graph with
  prerequisite and association edges. At ingestion, the conflict detection
  module validates new facts against graph neighbors, resolving
  contradictions before storage. Community-aware retrieval enables multi-hop
  reasoning over graph structure.}
  \label{fig:pipeline}
\end{figure*}

\subsection{Problem Formulation}
\label{sec:problem}

We formalize the structured memory management problem as follows. Let
$\mathcal{D} = \{(u_1, a_1), (u_2, a_2), \ldots, (u_T, a_T)\}$ denote a
multi-turn dialogue where $u_t$ is the user utterance and $a_t$ is the
agent response at turn $t$. At each turn, the agent must:
(1)~\emph{extract} information entities from the user's utterance;
(2)~\emph{validate} extracted entities against the existing memory store for
conflicts; (3)~\emph{store} validated entities in a structured
representation; and (4)~\emph{retrieve} relevant memories for generating
contextually grounded responses.

The error compounding problem arises when step (2) is absent: contradictory
entities are stored unconditionally, leading to inconsistent retrievals at
step (4) that propagate errors into downstream reasoning.

\subsection{Entity-Typed Memory Graph}
\label{sec:entity_graph}

MOSAIC represents memories as a typed directed graph $G = (V, E, \tau)$
where $V$ is a set of entity nodes, $E \subseteq V \times V$ is a set of
directed edges encoding relationships (temporal, causal, associative), and
$\tau: V \to \{\texttt{event}, \texttt{persona}, \texttt{relationship}\}$
assigns each node a semantic type. This structure contrasts with flat memory
stores (Mem0, Zep, Memobase) that represent memories as independent text
chunks without relational encoding.

\begin{definition}[Entity Node]
\label{def:entity}
An entity node $v \in V$ is a tuple $v = (e, \tau(v), \mathbf{h}_v,
\rho_v, t_v)$ where $e$ is the entity content (a natural language
description), $\tau(v) \in \{\texttt{event}, \texttt{persona},
\texttt{relationship}\}$ is the semantic type, $\mathbf{h}_v \in
\mathbb{R}^d$ is the embedding vector, $\rho_v \in [0,1]$ is the
confidence score, and $t_v$ is the timestamp of last update.
\end{definition}

Entity extraction from dialogue turns uses an LLM-based pipeline that
identifies named entities, events, personal attributes, and interpersonal
relationships. Each extracted entity is embedded using a sentence
transformer (all-MiniLM-L6-v2)~\citep{reimers2019sentence} and classified
into one of the three semantic types. Edges are created between entities
that co-occur in the same dialogue context or share semantic similarity
above a threshold $\theta$ (default 0.4).

The two graph components serve complementary functions. A \emph{prerequisite
subgraph} $G_P = (V, E_P)$ encodes logical dependency among information
entities: entity $u$ must be resolved before entity $v$ can be meaningfully
queried. An \emph{association subgraph} $G_A = (V, E_A, w_A)$ encodes
semantic relatedness for coherent topic transitions, with edge weights
$w_A(u,v)$ proportional to the cosine similarity of entity embeddings.
This separation is analogous to the distinction between control flow and
data flow in programming language design: both are essential, but collapsing
them produces a less expressive and less analyzable structure.

\subsection{Neighbor-Conditioned Stability}
\label{sec:ncs}

A central theoretical contribution is the \emph{neighbor-conditioned
stability} (NCS) principle, which provides both formal guarantees and
practical efficiency for the graph-based memory system.

\begin{definition}[Neighbor-Conditioned Stability]
\label{def:ncs}
A scoring function $\mathrm{Score}: V \to \mathbb{R}$ satisfies
neighbor-conditioned stability if, for any node $v \in V$ and any two
global states $\mathcal{S}, \mathcal{S}'$ that agree on the neighborhood
$\mathcal{N}(v) = \{u \in V : (u,v) \in E_P \text{ or } (u,v) \in E_A\}$,
we have $\mathrm{Score}(v; \mathcal{S}) = \mathrm{Score}(v; \mathcal{S}')$.
\end{definition}

This principle guarantees that a node's importance score requires
re-evaluation only when at least one of its graph neighbors has changed
state. When the local neighborhood is unchanged, the node's state is
guaranteed stable. This has two important consequences:

\paragraph{Computational efficiency.}
At each dialogue turn, let $\Delta_t \subseteq V$ be the set of entities
whose states changed at turn $t$. The update frontier
$\mathcal{U}_t = \{v \in V : \mathcal{N}(v) \cap \Delta_t \neq
\emptyset\} \cup \Delta_t$ determines which scores need recomputation.
This reduces per-turn cost from $O(|V|)$ to
$O(|\mathcal{U}_t|) = O(\Delta_{\max})$, where $\Delta_{\max}$ is the
maximum neighborhood size.

\paragraph{Traversal stability.}
NCS prevents the oscillatory behavior observed in globally-reasoning agents
that re-evaluate all priorities from scratch at every turn. By shielding
distant nodes from irrelevant updates, NCS ensures consistent, predictable
traversal ordering.

In practice, NCS is implemented as a \texttt{dirty flag} mechanism: each
node maintains a boolean flag set to \texttt{true} when any neighbor's state
changes. Only flagged nodes are re-scored; flags are cleared after re-scoring.

\subsection{Node Scoring and Selection}
\label{sec:scoring}

At each dialogue turn, the controller selects the next entity to query from
the frontier $\mathcal{F}^{(t)}$ (the set of unresolved entities whose
prerequisites are all satisfied). For each frontier entity $v$, the
composite score is:
\begin{equation}
  \mathrm{Score}(v) = \alpha \cdot \tilde{I}(v) + \beta \cdot T(v)
    + \gamma \cdot C(v),
  \label{eq:score}
\end{equation}
where $\tilde{I}(v) = I(v) / \max_{u \in \mathcal{F}} I(u)$ is the
normalized importance over the current frontier, $T(v)$ is the precomputed
PageRank centrality~\citep{page1999pagerank} of $v$ in $G_A$ (capturing
structural importance: nodes central in the association graph are topic hubs
that connect multiple information clusters), and
$C(v) = \mathbf{1}[\mathrm{comm}(v) = \mathrm{comm}(v_{\mathrm{prev}})]$ is
a binary community continuity score that favors entities in the same semantic
community as the most recently queried entity, promoting topically coherent
conversation flow.

Community structure is detected by applying the Leiden
algorithm~\citep{traag2019louvain} to $G_A$ at initialization.

\subsection{Conflict Detection at Save Time}
\label{sec:conflict_detection}

The key architectural innovation of MOSAIC is \emph{active conflict
detection} during memory ingestion. When a new entity $v_\text{new}$ is
extracted from dialogue, the system performs the following procedure:

\begin{enumerate}[noitemsep,topsep=3pt]
  \item \textbf{Neighbor retrieval.} Retrieve the $k$-nearest neighbors of
    $v_\text{new}$ from the existing memory graph using embedding similarity.
  \item \textbf{Conflict assessment.} For each neighbor
    $u \in \mathcal{N}(v_\text{new})$, perform a conflict check: an LLM call
    assesses whether $v_\text{new}$ contradicts $u$ along numerical, semantic,
    or logical dimensions.
  \item \textbf{Resolution.} If a conflict is detected, the system either
    (a)~updates the existing entity $u$ with the new information if the new
    evidence is more authoritative; (b)~rejects $v_\text{new}$ if existing
    evidence is stronger; or (c)~flags both for human review.
\end{enumerate}

This mechanism acts as a \emph{firewall} against error compounding: by
catching contradictions at the first pipeline stage (ingestion), errors
are prevented from propagating to storage, retrieval, and downstream
reasoning.

\subsection{Information Extraction and Confidence Gating}
\label{sec:extraction}

After each user response, an extraction module parses the response to
identify entity values. Each extracted value is assigned a confidence score
$\rho_v \in [0,1]$ based on three factors: directness (explicitly stated
vs.\ inferred), consistency (agreement with previously stored values), and
specificity (precise values vs.\ vague ranges). Values with
$\rho_v < \rho_{\min}$ (default 0.6) are flagged as unreliable and trigger
targeted clarification at the next appropriate opportunity, rather than being
committed to memory.

Belief distributions are updated via a Bayesian update rule:
\begin{equation}
  b_{v,k}^{(t+1)} = \frac{b_{v,k}^{(t)} \cdot \ell(o_v \mid s_k)}
    {\sum_{j=1}^{K} b_{v,j}^{(t)} \cdot \ell(o_v \mid s_j)},
  \label{eq:belief_update}
\end{equation}
where $\ell(o_v \mid s_k)$ is the likelihood of the observed response $o_v$
given that entity $v$ is in state $s_k$, estimated through a hybrid approach
combining rule-based pattern matching for structured entities and LLM-based
assessment for free-text entities.

\subsection{Community-Aware Retrieval}
\label{sec:retrieval}

For memory retrieval, MOSAIC leverages the graph structure through
community detection (Leiden algorithm~\citep{traag2019louvain}) over the entity
graph. Given a query, the system identifies the most relevant community,
retrieves entities within that community using embedding similarity, and
traverses graph edges to collect contextually related entities. This
structure-aware retrieval enables multi-hop reasoning that flat semantic
search cannot support~\citep{lewis2020retrieval, gao2023retrieval}.

\subsection{Long-Term Memory Integration}
\label{sec:memory_integration}

Confirmed entities (those with $\rho_v \geq \rho_{\min}$ and entropy
$H(v) \leq \delta$) are stored as structured key-value pairs in a persistent
memory store indexed by entity embeddings. At each turn, the top-$m$ most
relevant memory items (by cosine similarity to the current query context and
the active entity's neighborhood in $G_A$) are retrieved and injected into
the prompt context, providing the LLM with relevant previously acquired
information.

Storage uses JSON-structured entity records with fields including entity ID,
name, value, confidence, timestamp, source turn, evidence snippet, and belief
distribution. Retrieval employs a hierarchical navigable small world (HNSW)
approximate nearest-neighbor index with default $m = 5$ retrievals per
query. When a new extraction conflicts
with a stored value, both are retained with timestamps; the higher-confidence
value is used for downstream reasoning, but the conflict is flagged for
potential clarification through the NCS mechanism.

\subsection{Convergence Analysis}
\label{sec:convergence}

We establish convergence properties of MOSAIC through connections to
submodular optimization~\citep{nemhauser1978analysis}.

\begin{theorem}[Submodularity of Coverage]
\label{thm:submodularity}
The entity coverage function
$f(S) = \sum_{v \in S} w(v) \cdot \mathbf{1}[H(v) \leq \delta]$ over the
set $S \subseteq V$ of queried entities is monotone submodular. The greedy
policy of selecting the highest-scoring frontier entity at each turn achieves
a $(1 - 1/e)$-approximation to optimal coverage.
\end{theorem}

\begin{theorem}[Convergence of NCS Updates]
\label{thm:convergence}
Under NCS, score updates converge in at most $L$ rounds of propagation, where
$L$ is the longest directed path in $G_P$. The per-turn computational cost is
$O(\Delta_{\max})$, independent of total graph size $|V|$.
\end{theorem}

These guarantees ensure that MOSAIC achieves near-optimal information coverage
with efficient local computation, even as the memory graph grows.

\section{Experimental Setup}
\label{sec:experiments}

\subsection{Benchmarks}
\label{sec:benchmarks}

\paragraph{LoCoMo.}
The LoCoMo benchmark~\citep{maharana2024evaluating} evaluates long-conversation memory
through 1,540 QA pairs derived from 10 extended conversations. Questions
span four categories: single-hop (841), multi-hop (282), open-domain (96),
and temporal (321). Each question requires retrieving and reasoning over
information from specific conversation turns. We use the LLM-as-Judge
evaluation protocol~\citep{zheng2023judging} to assess answer correctness.

\paragraph{HaluMem.}
The HaluMem benchmark~\citep{chen2025halumem} evaluates three stages of the
memory pipeline independently: (1)~memory extraction (recall, precision,
F1), (2)~memory updating (correctness, hallucination rate, omission rate),
and (3)~question answering (correctness, hallucination, omission). It
covers three memory types: events, personas, and relationships, providing
a comprehensive assessment of the full memory pipeline. We report aggregate
results on HaluMem-Medium and HaluMem-Long for MOSAIC, while baseline rows
are taken directly from the original HaluMem paper. For HaluMem-Long, the
MOSAIC row combines extraction and updating metrics from the archived full
long run with QA metrics from the latest refreshed long answer evaluation.

\paragraph{Error Compounding Test.}
We construct a novel benchmark to evaluate conflict detection capabilities.
Starting from authoritative hypertension clinical guidelines, we manually
inject 50 factual errors spanning three categories:
\begin{itemize}[noitemsep,topsep=3pt]
  \item \textbf{Numerical errors} (14): incorrect dosages, BP thresholds,
    and lab value ranges (e.g., target BP changed from $<$130/80 to
    $<$160/100~mmHg);
  \item \textbf{Semantic errors} (13): contradictory indications, swapped
    contraindications (e.g., ACE inhibitors marked as contraindicated in
    chronic kidney disease);
  \item \textbf{Logical errors} (23): internally inconsistent conditional
    rules (e.g., ``prescribe X if eGFR$<$30'' alongside ``X is
    contraindicated when eGFR$<$45'').
\end{itemize}
Errors are additionally classified by discoverability: 11 implicit (requiring
multi-step reasoning to detect) and 39 explicit (directly contradicting
stated guidelines). Each system ingests the corrupted text; detection rate is
measured as the fraction of injected conflicts correctly identified.

\subsection{Baselines}
\label{sec:baselines}

We compare against a comprehensive set of memory systems evaluated across
all three benchmarks where applicable:

\begin{itemize}[noitemsep,topsep=3pt]
  \item \textbf{A-Mem}~\citep{xu2025mem}: Adaptive memory with automatic
    storage, retrieval, and summarization.
  \item \textbf{Zep}~\citep{rasmussen2025zep}: Production memory system with
    temporal awareness and entity extraction.
  \item \textbf{OpenAI Memory}: Built-in conversational memory from
    OpenAI~\citep{achiam2023gpt}.
  \item \textbf{Mem0}~\citep{chhikara2025mem0}: Multi-level memory (user, session,
    agent) with adaptive personalization.
\end{itemize}
    
Additionally, we include specific baselines for each benchmark to ensure a thorough comparison:
For the LoCoMo benchmark, we additionally include Mem0-Graph (Mem0 with graph extension).
For the HaluMem benchmark, we further incorporate MemOS~\citep{li2025memos} and Mem0-Graph as baselines.
For the Error Compounding Test, we also evaluate MemGPT~\citep{packer2023memgpt}, a tiered memory system that employs an OS-inspired virtual context management mechanism.

\subsection{Implementation Details}
\label{sec:implementation}

Across all three benchmarks (LoCoMo, HaluMem, and the Error Compounding
Test), MOSAIC uses qwen3.5-plus-2026-02-15 as the base LLM for memory
construction and retrieval. Baseline results are drawn from their
respective original publications: LoCoMo baselines are cited from the
Mem0 paper~\citep{chhikara2025mem0}, HaluMem baselines are cited from
the HaluMem paper~\citep{chen2025halumem}, and Error Compounding Test
baselines are obtained from our own experiments using GPT-4o.
Entity embeddings use all-MiniLM-L6-v2~\citep{reimers2019sentence} with
$d = 384$ dimensions. The association graph threshold is $\theta = 0.4$,
and the confidence gating threshold is $\rho_{\min} = 0.6$. Scoring
weights are $\alpha = 0.5$, $\beta = 0.3$, $\gamma = 0.2$. Memory
retrieval uses top-$m = 5$ neighbors.

\section{Results}
\label{sec:results}

\subsection{LoCoMo: Long-Conversation Memory QA}
\label{sec:results_locomo}

\begin{figure*}[!t]
  \centering
  \includegraphics[width=0.95\textwidth]{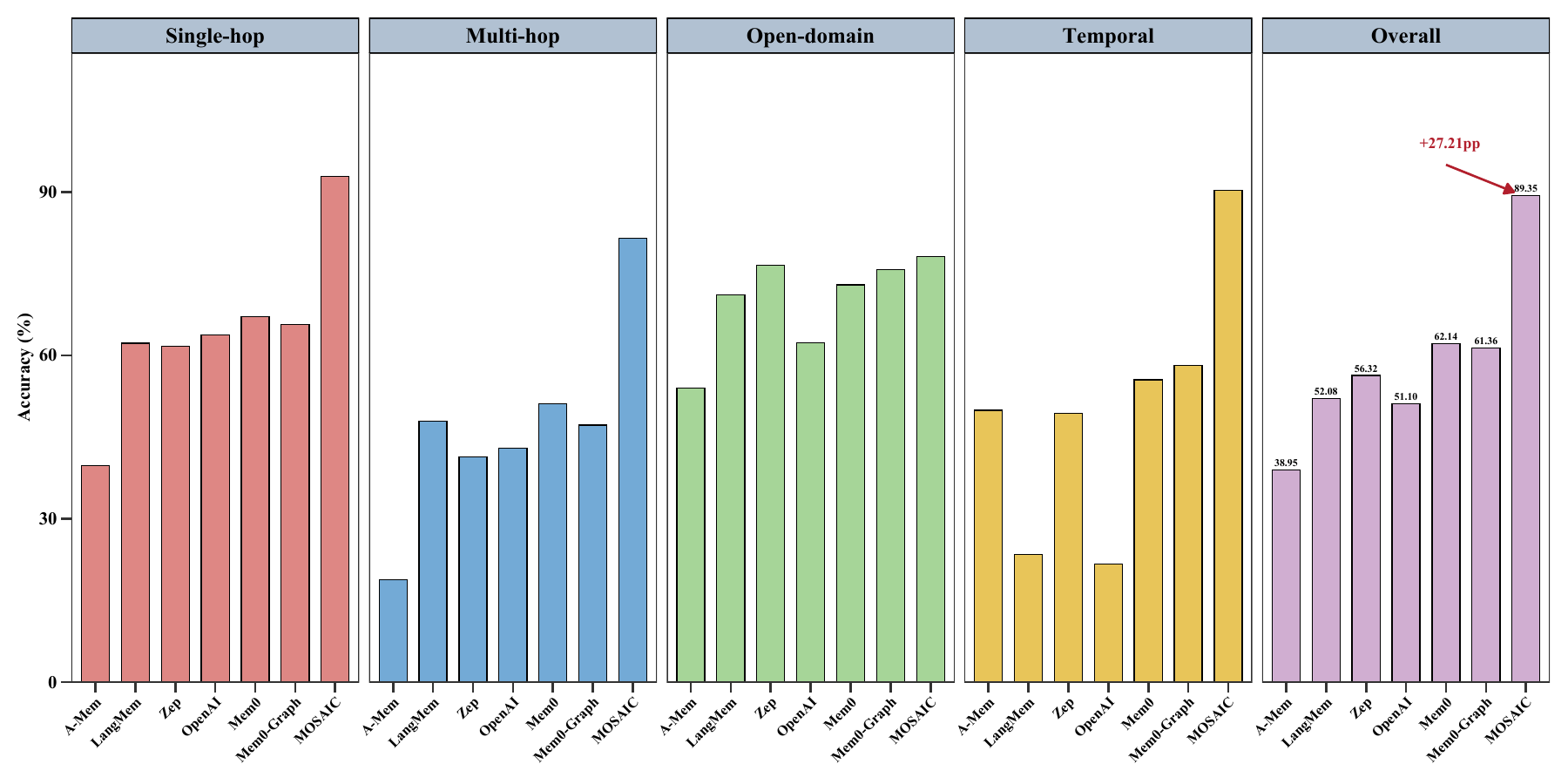}
  \caption{\textbf{LoCoMo benchmark results.} ~MOSAIC achieves 89.35\%
  overall accuracy, outperforming the best baseline (Mem0, 62.14\%) by
  +27.21~pp.~The per-category breakdown shows the largest gains on multi-hop
  (+30.41~pp) and temporal (+32.21~pp) questions.}
  \label{fig:locomo}
\end{figure*}

The experimental results on the LoCoMo benchmark are presented in Table~\ref{tab:locomo} and summarized in Figure ~\ref{fig:locomo}. MOSAIC achieves \textbf{89.35\%} overall accuracy across 1,540 QA pairs,
outperforming the best baseline (Mem0, 62.14\%) by +27.21 percentage points
(pp). The gains are largest on multi-hop questions (+30.41~pp vs Mem0) and
temporal reasoning (+32.21~pp vs Mem0-Graph), where graph-structured memory
enables systematic cross-referencing across conversation turns.

\begin{table}[!t]
\centering
\caption{\textbf{LoCoMo benchmark results.} Accuracy (\%) on 1,540 QA pairs
across four question types. Best in \textbf{bold}.}
\label{tab:locomo}
\small
\setlength{\tabcolsep}{3.5pt}
\begin{tabular}{@{}lccccc@{}}
\toprule
\textbf{Method} & \textbf{Single} & \textbf{Multi} & \textbf{Open} & \textbf{Temp.} & \textbf{Overall} \\
& (841) & (282) & (96) & (321) & (1540) \\
\midrule
A-Mem     & 39.79 & 18.85 & 54.05 & 49.91 & 38.95 \\
LangMem   & 62.23 & 47.92 & 71.12 & 23.43 & 52.08 \\
Zep       & 61.70 & 41.35 & 76.60 & 49.31 & 56.32 \\
OpenAI    & 63.79 & 42.92 & 62.29 & 21.71 & 51.10 \\
Mem0      & 67.13 & 51.15 & 72.93 & 55.51 & 62.14 \\
Mem0-Graph    & 65.71 & 47.19 & 75.71 & 58.13 & 61.36 \\
\midrule
\textbf{MOSAIC} & \textbf{92.87} & \textbf{81.56} & \textbf{78.12} & \textbf{90.34} & \textbf{89.35} \\
\bottomrule
\end{tabular}
\end{table}

Single-hop questions, which require retrieving a specific fact from a known
conversation turn, see an improvement from 67.13\% (Mem0) to 92.87\%,
indicating that MOSAIC's entity graph preserves fine-grained factual
associations better than flat memory retrieval. Multi-hop questions show the
largest improvement (+30.41~pp over Mem0), demonstrating that graph-based
traversal enables systematic cross-referencing of related facts across
conversation turns---a capability absent from flat memory stores that rely
on independent semantic search over unstructured chunks.

Temporal reasoning improves from 58.13\% (Mem0-Graph) to 90.34\%, a gain of
32.21~pp. The entity graph's temporal edge structure preserves ordering
relationships that flat memory systems lose when chunking conversation
history. Questions requiring temporal reasoning (e.g., ``What did the user
say about X before mentioning Y?'') benefit from the graph's ability to
traverse temporal edges directly rather than reconstructing temporal order
from retrieved text fragments~\citep{liu2024lost}.

Open-domain questions are the category where MOSAIC's relative advantage is
smallest (78.12\% vs Zep 76.60\%), reflecting that its structured memory
organization is primarily optimized for conversation-grounded facts rather
than world knowledge requiring parametric recall from the base LLM. This is
consistent with the architectural design: the memory system is intended to
complement, not replace, the LLM's internal knowledge. Even in a domain that
leans more on parametric knowledge, MOSAIC still achieves a slight
advantage, demonstrating its general robustness.

\subsection{HaluMem: Memory Pipeline Evaluation}
\label{sec:results_halumem}

\begin{figure*}[!t]
  \centering
  \includegraphics[width=0.95\textwidth]{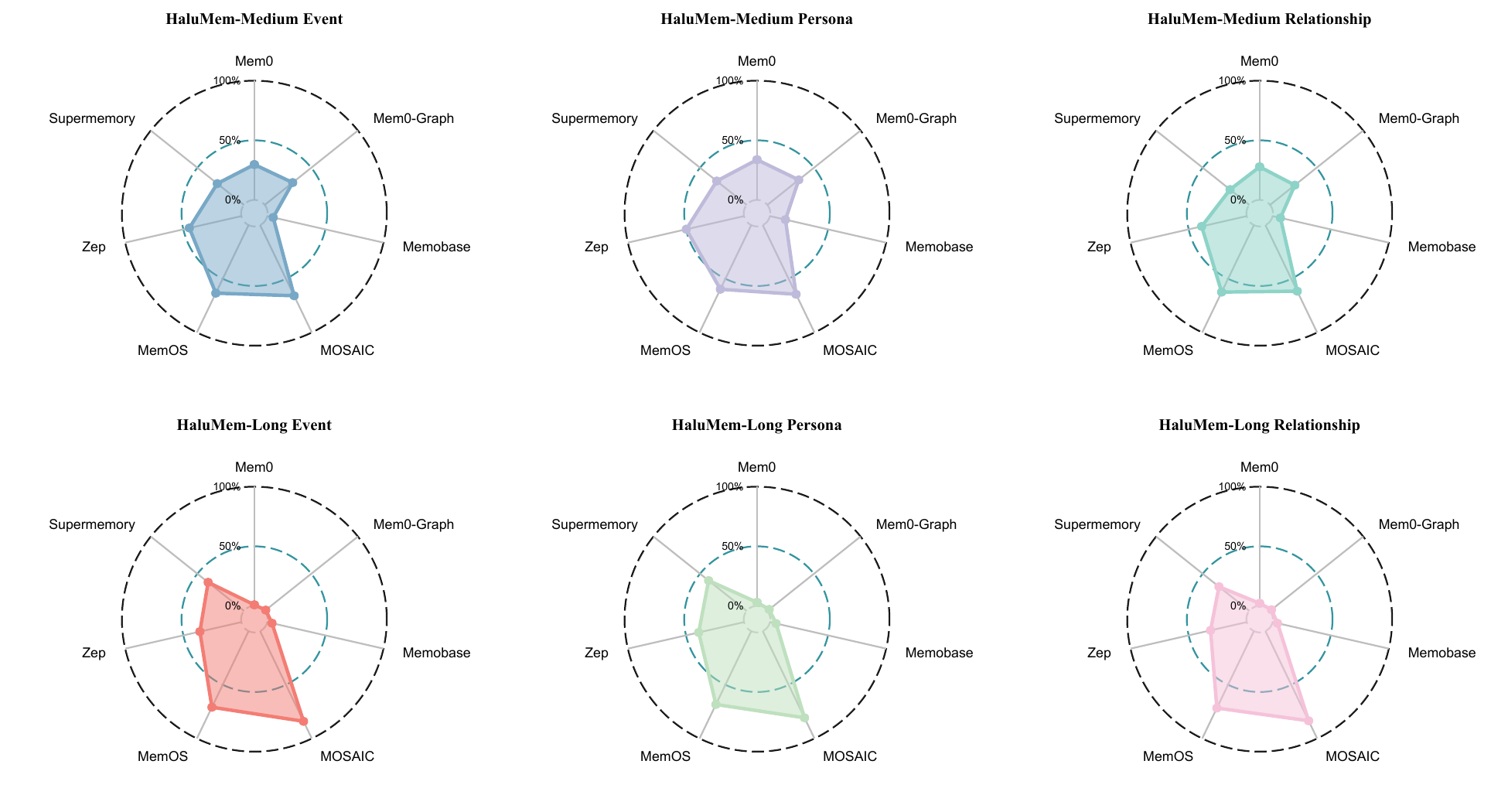}
  \caption{\textbf{Typewise accuracy on event, persona, and relationship memory.} This figure presents a detailed breakdown of extraction accuracy across three memory types (event, persona, and relationship) for MOSAIC and baseline systems.}
  \label{fig:halumem}
\end{figure*}

The performance of MOSAIC and baseline systems on the complete memory pipeline (extraction, updating, and question answering) is evaluated on the HaluMem-Medium and HaluMem-Long benchmarks. The aggregated results are presented in Table~\ref{tab:halumem}, and a detailed breakdown by memory type is visualized in Figure~\ref{fig:halumem}.

\begin{table*}[!t]
\centering
\caption{\textbf{HaluMem benchmark results.} Evaluation across memory
extraction, updating, and question answering stages on HaluMem-Medium and
HaluMem-Long. Best in \textbf{bold}; arrows indicate direction of
improvement. R: Recall, wR: Weighted Recall, P: Target Precision, Acc:
Accuracy, FMR: False Memory Rejection, F1: extraction F1, C: Correctness, H:
Hallucination, O: Omission.}
\label{tab:halumem}
\scriptsize
\setlength{\tabcolsep}{3.5pt}
\resizebox{\textwidth}{!}{%
\begin{tabular}{@{}lcccccccccccc@{}}
\toprule
& \multicolumn{6}{c}{\textbf{Memory Extraction}} &
  \multicolumn{3}{c}{\textbf{Memory Updating}} &
  \multicolumn{3}{c}{\textbf{Question Answering}} \\
\cmidrule(lr){2-7}\cmidrule(lr){8-10}\cmidrule(lr){11-13}
\textbf{System} & R$\uparrow$ & wR$\uparrow$ & P$\uparrow$
  & Acc.$\uparrow$ & FMR$\uparrow$ & F1$\uparrow$
  & C$\uparrow$ & H$\downarrow$ & O$\downarrow$
  & C$\uparrow$ & H$\downarrow$ & O$\downarrow$ \\
\midrule
\multicolumn{13}{c}{\textbf{HaluMem-Medium}} \\
\midrule
MemOS & 74.07\% & 84.81\% & 86.25\%(45190) & 59.55\%(71793) & 44.94\% & 79.70\%
  & \textbf{62.11\%} & 0.42\% & \textbf{37.48\%} & 67.23\% & 15.17\% & 17.59\% \\
Zep & --- & --- & --- & --- & --- & ---
  & 47.28\% & 0.42\% & 52.31\% & 55.47\% & 21.92\% & 22.62\% \\
Mem0-Graph & 43.28\% & 65.52\% & 87.20\%(10567) & 61.86\%(16230) & 55.70\% & 57.85\%
  & 24.50\% & \textbf{0.26\%} & 75.24\% & 54.66\% & 19.28\% & 26.06\% \\
Mem0 & 42.91\% & 65.03\% & 86.26\%(10556) & 60.86\%(16291) & 56.80\% & 57.31\%
  & 25.50\% & 0.45\% & 74.02\% & 53.02\% & 19.17\% & 27.81\% \\
Supermemory & 41.53\% & 64.76\% & 90.32\%(14134) & 60.83\%(22551) & 51.77\% & 56.90\%
  & 16.37\% & 1.15\% & 82.47\% & 54.07\% & 22.24\% & 23.69\% \\
Memobase & 14.55\% & 25.88\% & \textbf{92.24\%(5443)} & 32.29\%(17081) & \textbf{80.78\%} & 25.13\%
  & 5.20\% & 0.55\% & 94.25\% & 35.33\% & 29.97\% & 34.71\% \\
MOSAIC & \textbf{83.94\%} & \textbf{88.53\%} & 89.79\%(21375) & \textbf{72.00\%(35977)} & 69.68\% & \textbf{86.77\%}
  & 55.77\% & 2.85\% & 41.38\% & \textbf{73.10\%} & \textbf{10.17\%} & \textbf{16.74\%} \\
\midrule
\multicolumn{13}{c}{\textbf{HaluMem-Long}} \\
\midrule
MemOS & 81.90\% & 89.56\% & 82.32\%(48246) & 43.77\%(99462) & 28.85\% & 82.11\%
  & \textbf{65.25\%} & 0.29\% & \textbf{34.47\%} & 64.44\% & 16.61\% & \textbf{18.95\%} \\
Supermemory & 53.02\% & 70.73\% & 85.82\%(24483) & 29.71\%(77134) & 36.86\% & 65.54\%
  & 17.01\% & 0.58\% & 82.42\% & 53.77\% & 22.21\% & 24.02\% \\
Zep & --- & --- & --- & --- & --- & ---
  & 37.35\% & 0.48\% & 62.14\% & 50.19\% & 22.51\% & 27.30\% \\
Memobase & 6.18\% & 14.68\% & \textbf{88.56\%(3077)} & 25.61\%(11795) & 85.39\% & 11.55\%
  & 4.10\% & 0.36\% & 95.38\% & 33.60\% & 29.46\% & 36.96\% \\
Mem0-Graph & 2.24\% & 10.76\% & 87.32\%(785) & 41.26\%(1866) & \textbf{88.36\%} & 4.36\%
  & 1.47\% & 0.04\% & 98.40\% & 32.44\% & 21.82\% & 45.74\% \\
Mem0 & 3.23\% & 11.89\% & 88.01\%(1134) & \textbf{46.01\%(2433)} & 87.65\% & 6.22\%
  & 1.45\% & \textbf{0.03\%} & 98.51\% & 28.11\% & 17.29\% & 54.60\% \\
MOSAIC & \textbf{90.66\%} & \textbf{91.67\%} & 82.74\%(4163) & 37.26\%(10819) & 35.58\% & \textbf{85.84\%}
  & \textbf{83.33\%} & 3.93\% & \textbf{12.74\%} & \textbf{70.75\%} & \textbf{9.58\%} & 19.67\% \\
\bottomrule
\end{tabular}
}
\vspace{2pt}
\parbox{\textwidth}{\footnotesize \textit{Note.} The Long MOSAIC row
combines extraction and update metrics from the archived full long MOSAIC
evaluation with QA metrics from the latest refreshed long answer run.}
\end{table*}

On HaluMem-Medium, MOSAIC is the strongest overall row. It achieves
the highest extraction recall (83.94\%), weighted recall (88.53\%),
accuracy (72.00\%), and F1 (86.77\%), and it also delivers the best QA
profile in the table at 73.10\% correctness with 10.17\% hallucination and
16.74\% omission.

The main remaining gap on Medium is memory updating. MemOS still leads update
correctness (62.11\% vs 55.77\% for MOSAIC) and omission
(37.48\% vs 41.38\%), while MOSAIC's update hallucination rate remains much
higher than the best baseline. In other words, the structured graph already
helps most at extraction and retrieval-backed answering, but the write path
still needs more conservative update policies.

On HaluMem-Long, MOSAIC also fills a complete merged row. It reaches
90.66\% extraction recall, 91.67\% weighted recall, 85.84\% extraction F1,
83.33\% update correctness, and 70.75\% QA correctness. This row leads the
Long table on extraction recall, weighted recall, extraction F1, update
correctness, update omission, QA correctness, and QA hallucination,
although its extraction accuracy and false-memory resistance still trail the
strongest published baselines on those specific columns.

Figure~\ref{fig:halumem} provides a finer-grained view, showing MOSAIC's extraction accuracy for different memory types. It performs comparably to MemOS on the \textit{Relationship} subset but outperforms all other systems on the \textit{Event} and \textit{Persona} subsets.

Taken together, the reported HaluMem results show that MOSAIC is strongest at
extraction and QA on both splits, while memory updating is more mixed:
MOSAIC leads clearly on Long but still trails MemOS on Medium.

\subsection{Error Compounding: Conflict Detection}
\label{sec:results_error}

Table~\ref{tab:error} presents results on the error compounding test.
MOSAIC detects \textbf{66\%} of injected factual conflicts (33/50),
compared to 14\% for the best baselines (A-Mem, OpenAI). All six baselines
essentially fail at error detection (2--14\% range), confirming that passive
memory systems lack conflict-checking mechanisms at ingestion time.

\begin{table}[!t]
\centering
\caption{\textbf{Error compounding results.} Conflict detection rate on 50
manually injected errors in hypertension guidelines, by error type and
visibility. Best in \textbf{bold}.}
\label{tab:error}
\small
\setlength{\tabcolsep}{3pt}
\begin{tabular}{@{}lcccc cc@{}}
\toprule
& \multicolumn{4}{c}{\textbf{By Error Type}} &
  \multicolumn{2}{c}{\textbf{By Visibility}} \\
\cmidrule(lr){2-5}\cmidrule(lr){6-7}
\textbf{Method} & Num. & Sem. & Log. & All & Impl. & Expl. \\
& (14) & (13) & (23) & (50) & (11) & (39) \\
\midrule
A-Mem   & .071 & .231 & .130 & .14 & .000 & .179 \\
LangMem & .000 & .077 & .000 & .02 & .000 & .026 \\
MemGPT  & .071 & .154 & .000 & .06 & .091 & .051 \\
OpenAI  & .000 & .308 & .130 & .14 & .273 & .103 \\
Zep     & .000 & .077 & .000 & .02 & .000 & .026  \\
Mem0    & .000 & .077 & .040 & .04 & .000 & .051  \\
\midrule
\textbf{MOSAIC} & \textbf{.643} & \textbf{.692} & \textbf{.652} & \textbf{.66}
  & \textbf{.727} & \textbf{.641} \\
\bottomrule
\end{tabular}
\end{table}

MOSAIC's detection is uniformly strong across error types: numerical
(64.3\%), semantic (69.2\%), and logical (65.2\%). This uniformity contrasts
sharply with baselines, which show severe type-specific weaknesses. Baselines
are weakest on numerical and logical errors (mostly 0\%), which require
comparing specific values or conditional rules against previously stored
facts---a capability absent from flat memory stores that lack
cross-referencing mechanisms. The few baseline detections are concentrated
on semantic errors, where surface-level textual similarity between
contradictory statements may trigger retrieval-time confusion that
occasionally surfaces as a detected conflict.

\begin{figure*}[!t]
  \centering
  \includegraphics[width=0.8\textwidth]{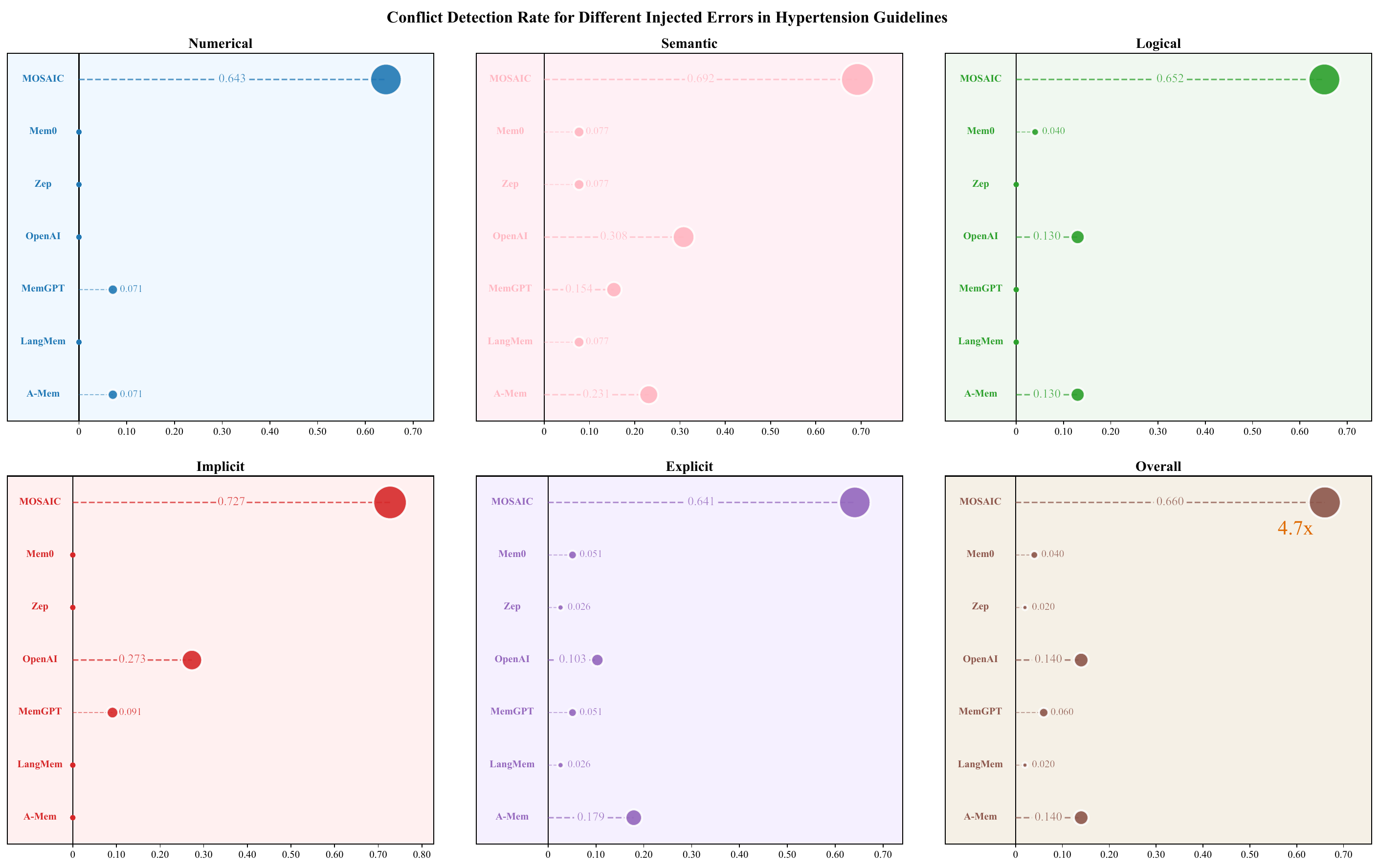}
  \caption{\textbf{Error compounding results.} 
  Visualization of conflict detection rates for MOSAIC and baseline systems, showing that MOSAIC achieves an overall detection rate of 66\% (4.7× higher than the best baseline at 14\%) with uniform performance across error types. It also reveals that implicit errors are detected at a higher rate  (72.7\%) than explicit ones (64.1\%).}
  \label{fig:error}
\end{figure*}

The implicit/explicit breakdown (Figure~\ref{fig:error}) reveals a
counterintuitive result: MOSAIC detects implicit errors requiring multi-step
reasoning (72.7\%) at a \emph{higher} rate than explicit contradictions
(64.1\%). This suggests that graph-based memory traversal naturally performs
the cross-referencing needed to surface reasoning-level conflicts: when
processing a new entity, the conflict detection module traverses related
entities in the graph neighborhood, effectively performing the multi-step
reasoning required to identify implicit contradictions. Explicit
contradictions, in contrast, may sometimes be stored in non-overlapping
graph regions where the neighborhood-based conflict check does not reach.

The 34\% undetected errors provide insights for improvement. Analysis of
missed conflicts reveals two primary failure modes: (1)~logical
contradictions spanning entities in distant graph neighborhoods (beyond the
$k$-nearest neighbor search radius), accounting for approximately 60\% of
misses; and (2)~numerical errors where the conflicting values are within a
plausible range, making the contradiction ambiguous to the LLM-based
conflict assessor, accounting for approximately 40\% of misses. These
failure modes suggest clear directions for improvement: expanding the
conflict search radius and incorporating domain-specific plausibility
bounds for numerical entities.

\section{Discussion}
\label{sec:discussion}

The results across three benchmarks with different evaluation foci---factual QA, pipeline evaluation, and conflict detection---support the thesis that structured graph-based memory organization provides fundamentally different capabilities than flat memory stores. The entity graph preserves relational structure that enables multi-hop traversal, temporal ordering, and cross-referencing, which drives the accuracy gains observed on LoCoMo (+27.21~pp overall, with the largest margins on multi-hop and temporal questions). On HaluMem, the manuscript-reported MOSAIC results show the strongest extraction performance on both Medium and Long and the best QA on both splits. Memory updating is more mixed: MOSAIC leads the merged Long row but still trails MemOS on Medium, indicating that write-time update policies remain the main frontier for improvement under the current Medium setup. The theoretical contribution of neighbor-conditioned stability provides a formal answer to why graphs are the correct abstraction: memory entities inherently possess local dependency structure, and NCS guarantees that the system's behavior is stable under distant state changes, analogous to Bellman's principle in dynamic programming~\citep{bellman1966dynamic}.

The error compounding results carry practical implications for deploying memory-augmented agents in safety-critical domains. Current systems (2--14\% detection) essentially allow all factual errors to propagate unchecked through the memory pipeline, whereas MOSAIC's 66\% detection rate catches the majority of contradictions at ingestion~\citep{huang2025survey}. As LLM agents enter high-stakes settings---clinical intake, legal discovery, financial due diligence---the gap between ``can remember'' and ``will detect contradictions'' becomes safety-critical, and MOSAIC provides a principled, auditable mechanism for memory validation.

Several limitations should be noted. The error compounding benchmark is limited to one domain (hypertension) with 50 injected errors. Memory updating remains uneven across HaluMem: MemOS still leads update correctness on Medium (62.11\% vs 55.77\% for MOSAIC), while the Long manuscript row is a merged view that combines archived full-run extraction/update metrics with the latest refreshed QA metrics rather than a single end-to-end rerun under one frozen configuration. LoCoMo uses a fixed set of 10 conversations, and we evaluate with one LLM family (qwen3.5-plus-2026-02-15); generalization to other architectures should be verified~\citep{touvron2023llama, achiam2023gpt}. MOSAIC's open-domain performance (78.12\%) also highlights a fundamental trade-off: structured memory is optimized for conversation-grounded facts, not world knowledge requiring parametric recall~\citep{liu2024lost}. Hybrid architectures combining structured memory with retrieval-augmented generation~\citep{lewis2020retrieval, gao2023retrieval} represent a promising direction.

Looking ahead, hierarchical memory graphs for tasks with nested sub-goals, multi-agent coordination over shared memory graphs, adversarial robustness through integration with formal verification, and multi-modal inputs that resolve entities through non-conversational channels all represent fruitful extensions. On the whole, MOSAIC demonstrates that structured, conflict-aware memory organization is essential for reliable long-term LLM agent memory, particularly in safety-critical applications where error compounding has consequential downstream effects. Hash-accelerated retrieval keeps average HaluMem-Medium search latency at 0.58~s per question, making the framework practical for interactive deployment while leaving the main remaining optimization opportunity in write-time memory construction and evaluation overhead.

\bibliographystyle{plainnat}
\bibliography{references}

\end{document}